\documentclass[a4paper,11pt]{article}
\usepackage{amsfonts}
\usepackage{amsmath}% AMSLaTeXºê°ü ÓÃÀ´Åųö¸ü¼ÓƯÁÁµÄ¹«Ê½
\usepackage{amssymb}% AMSLaTeXºê°ü ÓÃÀ´Åųö¸ü¼ÓƯÁÁµÄ¹«Ê½
\usepackage{amsthm}%¶¨ÀíºÍÖ¤Ã÷»·¾³
\usepackage{color,bibcheck}
\usepackage{algorithm,algorithmic}
\allowdisplaybreaks[4]%¹ÄÀø¹«Ê½×Ô¶¯·ÖÒ³
\usepackage{mathrsfs, graphicx,hyperref}
\usepackage[all]{xy}
\usepackage{amsmath}
\UseRawInputEncoding
\begin{document}
\title{CTG-Net: An Efficient Cascaded Framework Driven by Terminal Guidance Mechanism for Dilated Pancreatic Duct Segmentation}

\author{Liwen Zou\footnote{E-mail: dz20210008@smail.nju.edu.cn. Department of Mathematics, Nanjing University, Nanjing 210008, P.R. China.}, Zhenghua Cai, Yudong Qiu, Luying Gui, Liang Mao\footnote{(Corresponding author) E-mail: maoliang@njglyy.com. Department of General Surgery, Nanjing Drum Tower Hospital, Nanjing 210008, P.R. China.} and Xiaoping Yang\footnote{(Corresponding author) E-mail: xpyang@nju.edu.cn. Department of Mathematics, Nanjing University, Nanjing 210008, P.R. China.}}

\maketitle

\begin{abstract}
Pancreatic duct dilation indicates a high risk of various pancreatic diseases. Segmentation of dilated pancreatic ducts on computed tomography (CT) images shows the potential to assist the early diagnosis, surgical planning and prognosis. Because of the ducts' tiny sizes, slender tubular structures and the surrounding distractions, most current researches on pancreatic duct segmentation achieve low accuracy and always have segmentation errors on the terminal parts of the ducts. To address these problems, we propose a terminal guidance mechanism called cascaded terminal guidance network (CTG-Net). Firstly, a terminal attention mechanism is established on the skeletons extracted from the coarse predictions. Then, to get fine terminal segmentation, a subnetwork is designed for jointly learning the local intensity from the original images, feature cues from coarse predictions and global anatomy information from the pancreas distance transform maps. Finally, a terminal distraction attention module which explicitly learns the distribution of the terminal distraction is proposed to reduce the false positive and false negative predictions. We also propose a new metric called tDice to measure the terminal segmentation accuracy for targets with tubular structures and two segmentation metrics for distractions. We collect our dilated pancreatic duct segmentation dataset with 150 CT scans from patients with 5 types of pancreatic tumors. Experimental results on our dataset show that our proposed approach boosts dilated pancreatic duct segmentation accuracy by nearly 20\% compared with the existing results, and achieves more than 9\% improvement for the terminal segmentation accuracy compared with the state-of-the-art methods.


\end{abstract}

\section{Introduction}
\begin{figure}[h]
		\centering
		{
			\includegraphics[width=12cm,]{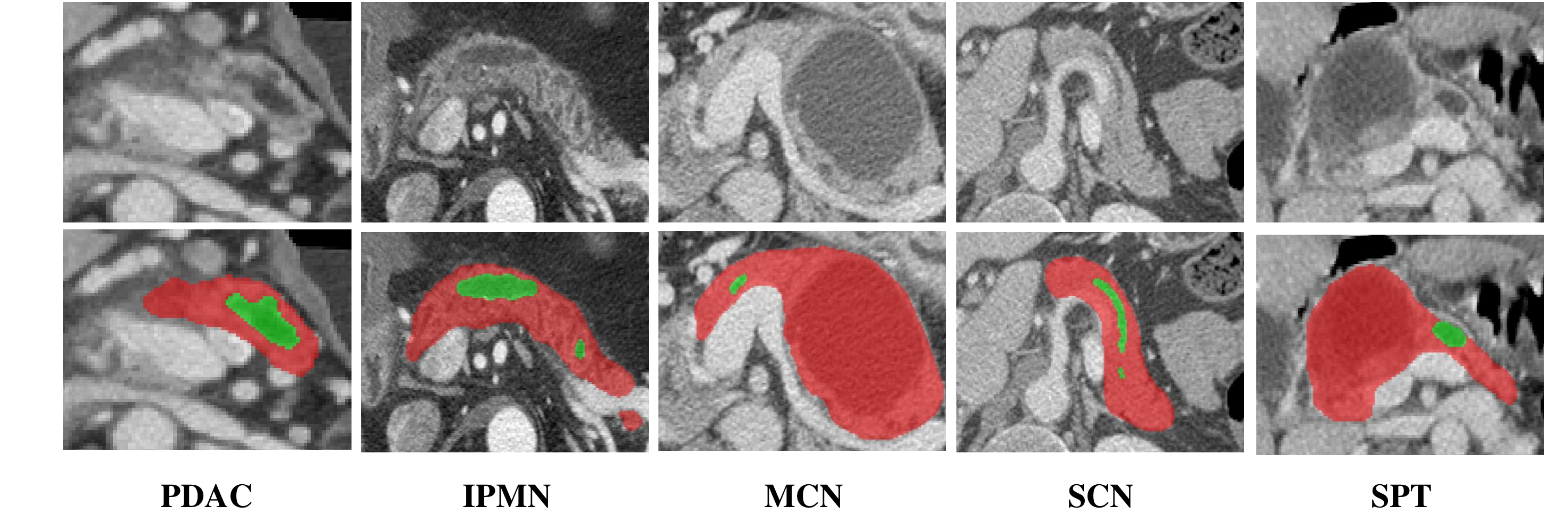}
		}
		\caption{A comparison of dilated pancreatic ducts with 5 different pancreatic tumors. \textbf{Top:} the CT images. \textbf{Bottom:} the corresponding masks where red denotes the pancreas and green denotes the dilated pancreatic duct.}\label{mass}
	\end{figure}

Pancreatic cancer is one of the most intractable cancers with only 10\% of 5-year relative survival rate in the USA. The most common pancreatic cancer occurs in the main pancreatic duct, known as pancreatic ductal adenocarcinoma (PDAC)\cite{Mizrahi}. Pancreatic duct dilation is identified as a high risk of PDAC in several clinical studies\cite{Edge,Tanaka}. However, the pancreatic disease taxonomy related to dilated pancreatic duct includes other types of tumors such as intraductal papillary mucinous neoplasm (IPMN), serous cystic intraductal papillary mucinous neoplasm (SCN), intraductal papillary mucinous neoplasm (MCN) and solid pseudopapillary tumor (SPT)\cite{Simeon}. A comparison of pancreatic duct dilations with 5 different pancreatic tumors on computed tomography (CT) images are shown in Figure \ref{mass}. Because of the slender shape and tiny size of the pancreatic duct, the normal pancreatic duct region is almost invisible on contrast-enhanced abdominal CT. Visibility of pancreatic duct from CT image could be a warning sign for PDAC and other pancreatic tumors. Hence, automated segmentation for dilated pancreatic ducts on CT images is promising for the early diagnosis, surgical planning and prognosis of pancreatic tumors.

\begin{figure}[h]
		\centering
		{
			\includegraphics[width=12cm,]{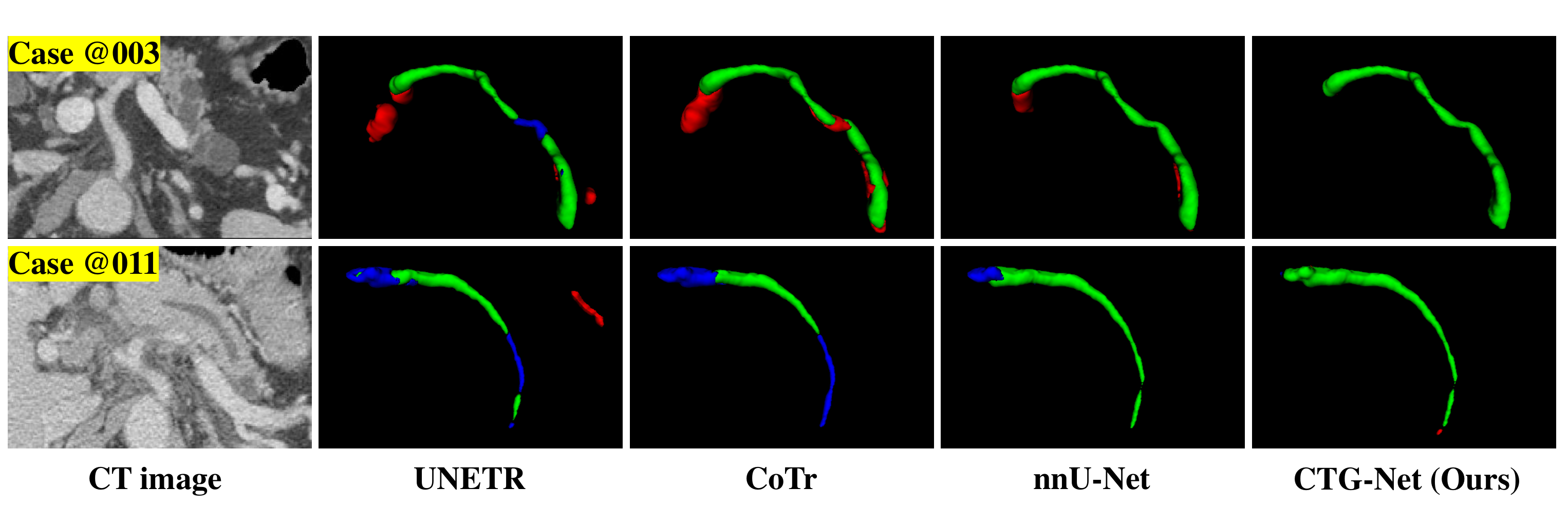}
		}
		\caption{Visual comparison of the dilated pancreatic duct segmentation. The first column shows input CT cases in our experiments, and the second to last columns show the segmentation results by different networks with cascaded strategy and our proposed method. The green, red and blue voxels denote the true positive, false positive and false negative regions.}\label{terminal}
	\end{figure}

The pancreas has a small size in the abdomen CT scan, therefore, a pancreatic duct inside the pancreas has a much smaller size, which makes the segmentation quite difficult. Additionally, the intensity similarity, slender shape, and distractions from surrounding tissues and lesions usually lead to the false positive (FP) or false negative (FN) predictions outside the pancreas or on the terminal regions of the duct predictions. The general experience motivating the solution is that when radiologists look for the pancreatic ducts, they usually concentrate on the pancreas regions and ignore other regions' contexts. This suggests that a deep neural network should also mainly concentrate on the pancreas when it is designed for pancreatic duct segmentation. Therefore, the cascaded strategy of segmenting pancreas firstly and pancreatic duct secondly shows better performance than segmenting the duct on the whole CT scans \cite{Shen}. However, previous methods for dilated pancreatic duct segmentation still can not achieve satisfactory performance mostly because of the ducts' slender shape and confusing tissues such as bile ducts and tumors around the pancreatic ducts. In fact, since the pancreatic ducts have tubular structures, almost all of the errors are located in the terminal regions of the predictions. Figure \ref{terminal} shows the experimental results that we compare the cascaded convolutional neural network (CNN)-based methods and our proposed method, where we can see that general cascaded methods can not reduce the terminal errors well.

To address this problem, we propose a new cascaded terminal guidance network (CTG-Net) to explicitly take the terminal guidance for segmentation into account. Our CTG-Net includes 3 stages: coarse, fine and refine stage. The coarse stage uses the traditional cascaded framework composed of a pancreas segmentation module (PSM) and a pancreatic duct segmentation module (PDSM) to segment the pancreas firstly and pancreatic duct secondly. The coarse pancreas and duct predictions are concatenated with the CT images cropped by pancreas regions of interest (RoI) to be fed into the fine stage with our proposed terminal guidance segmentation module (TGSM). To focus on the terminal errors of the coarse predictions, our TGSM firstly skeletonizes the coarse duct masks to get the centerlines and builds graphs by connected points on these centerlines. Secondly, all of the endpoints in these graphs will be find by neighbor selection to extract the terminal regions of the coarse duct predictions. Considering the limited local information on these terminal regions, we adopt the pancreas distance transform map as the global anatomy guidance to improve the terminal segmentation. Then, a three-channel U-shape network is designed for jointly learning the local intensity from the terminal duct CT images, feature cues from the coarse predictions and global anatomy information from the pancreas distance transform maps. In the refine stage, we propose a terminal distraction attention module (TDAM) to learn the terminal distraction information and distinguish the false positive and false negative regions as possible to improve the predictions obtained from the fine stage. We collect our dilated pancreatic duct segmentation (DPDS) dataset with 150 CT scans from patients with 5 types of pancreatic tumors. Experimental results on our dataset show the superiority to the previous techniques especially for the terminal segmentation accuracy. Moreover, we propose a new metric, called terminal Dice (tDice), to measure the terminal segmentation accuracy for targets with tubular structures, and two segmentation metrics for distractions.

The main contributions of our work are summarized as follows.
\begin{itemize}
\item To the best of our knowledge, this is the first work to consider the terminal errors caused by traditional CNN-based segmentation methods for objects with tubular structures, and we propose a new cascaded framework driven by terminal guidance mechanism for the dilated pancreatic duct segmentation to solve the problem.

\item We design a terminal guidance segmentation module (TGSM), which focuses on the terminal predictions and takes the local intensity, global anatomy information and the feature cues of the targets into account to improve the terminal segmentation accuracy for dilated pancreatic ducts.

\item We design a terminal distraction attention module (TDAM) that explicitly learns the distribution of the distraction tissues within the terminal regions to reduce the terminal segmentation errors.

\item We propose a new metric called tDice to measure the terminal segmentation accuracy for targets with tubular structures and two segmentation metrics for distractions.

\item Our proposed DPDS dataset includes not only the PDAC patients but also the cases with other 4 types of pancreatic tumors. Experimental results on our dataset show that our proposed approach boosts dilated pancreatic duct segmentation accuracy by nearly 20\% compared with the existing results, and achieves more than 9\% improvement for the terminal segmentation accuracy compared with the state-of-the-art (SOTA) methods.
\end{itemize}

\section{Related work}
\subsection{Deep learning based methods for medical image analysis}
Deep-learning-base methods have successfully solved many challenging tasks in image processing, such as classification \cite{resnet}, semantic segmentation \cite{lgac,abdct} and object detection \cite{rw}. With the rapid development of the fully convolutional network (FCN)\cite{FCN}, its extensions \cite{unet,vnet,AttentionU-net, nnunet} are widely applied in the segmentation task during medical image analysis. Encoder and decoder structures greatly improve the performance on pixel-wise segmentation compared with many traditional methods which need designing manual features. Semantic segmentation for medical images is important for developing computer-aided diagnosis (CAD) and computer-assisted surgery (CAS) systems.

\subsection{Previous techniques for dilated pancreatic duct segmentation}
Only few studies are related to pancreatic duct segmentation\cite{Zhou,Xia,Shen,ShenC}. Zhou \textit{et al.}\cite{Zhou} investigated a dual-path network for pancreas, PDAC tumor, and pancreatic duct segmentation. Both arterial and venous phase CT volumes are required to conduct the segmentation processes. They achieved a Dice socre of 56.77\% in their collected PDAC dataset including 239 cases with dilated pancreatic ducts. Xia \textit{et al.}\cite{Xia} proposed a segmentation method based on multi-phase CT alignment that achieved better performance of 64.38\% Dice sore on the same dataset. In \cite{Shen}, they used a cascaded segmentation framework with a pancreas segmentation network to get the pancreas mask RoI and segment the pancreatic duct with the following U-like network with bottleneck attention block. The got 49.87\% Dice score on their collected PDAC dataset including 30 cases with dilated pancreatic ducts. Shen \textit{et al.}\cite{ShenC} embedded pancreas-guided attention mechanism to the segmentation network and improved the segmentation accuracy of \cite{Shen} to 54.16\% Dice score on same single-phase CT dataset. Almost all of the previous techniques have not pay attention to the terminal segmentation errors caused by the tiny size and slender shape of the dilated pancreatic ducts.  Actually, the low segmentation accuracy can not meet the clinical needs well. Moreover, almost all of these researches are focused on PDAC patients, the segmentation for dilated pancreatic duct cases with other pancreatic tumors needs to be addressed.

\subsection{Cascaded strategy for medical image segmentation}
Multiple applications of cascaded methods are used for medical image segmentation. Chang and Teng presented a two-stage self-organizing map approach, which can precisely identify dominant color components to discover the region of interest for diagnosis purposes \cite{Chang}. Roth \textit{et al.} claimed a two-stage, coarse-to-fine approach for three anatomical organ segmentation (liver, spleen, pancreas). They first used a 3D FCN to roughly define a candidate region and then fed the region to the second 3D FCN, which focuses on more detailed segmentation of the organs and vessels \cite{Roth}. For segmentation of lesions within organs, the cascaded strategy is an effective framework to create an attention mechanism to guide the model where it should focus.

\subsection{Distance transform for tubular structure}
Distance transform \cite{distancetransform} is a classical image processing operator to produce a distance map with the same size of the input image, each value on the pixel/voxel is the distance from the foreground pixel/voxel to the foreground boundary. Distance transform is also known as the basis of one type of skeletonization algorithms \cite{Ge}. Thus, the distance map encodes the geometric characteristics of the tubular structure. In \cite{deepdistance}, the authors proposed a geometry-aware tubular structure segmentation method which combines intuitions from the classical distance transform for skeletonization and modern deep segmentation networks. In \cite{lgac}, the level set function (also as one distance transform) was used to measure the segmentation errors and a new loss function was designed based on it to train their segmentation network. Inspired by these works above, we observe that the distance transform map of the pancreas can reflect the anatomy structure of the pancreas itself. Every value on the voxel shows the distance to the boundary of the pancreas, which can provide more anatomy details inside the pancreas for the dilated pancreatic duct segmentation.

\subsection{Distraction attention}
Distraction concepts have been explored in many computer vision tasks, such as semantic segmentation \cite{Huang}, saliency detection \cite{Chen,Xiao} and visual tracking \cite{Zhu}. Most of previous methods suppress negative high-level representations or filter out the distracting regions \cite{Xiao,Huang,Chen}. Instead, Zheng \textit{et al.} used distraction cues to improve the performance of shadow distraction \cite{Zheng}. They split shadow distraction into false negative estimates and false positive estimates, and designed specific architectures to efficiently integrate the two types of distraction semantics. In \cite{Zhao}, the authors proposed a cascaded two-stage U-Net model to explicitly take the ambiguous region information into account. In their model, the first stage generates a global segmentation for the whole input CT volume and predicts latent distraction regions, which contain both false negative areas and false positive areas, against the segmentation ground truth. The second stage embeds the distraction region information into local segmentation for volume patches to further discriminate the target regions. Inspired by this method, we develop a terminal distraction attention module to explicitly learn semantic features of the terminal distraction regions, and embed these features in U-Net structure for the terminal segmentation errors.

\section{Methods}
\subsection{Network architecture}
\begin{figure*}[htbp]
		\centering
		{
			\includegraphics[width=15cm,]{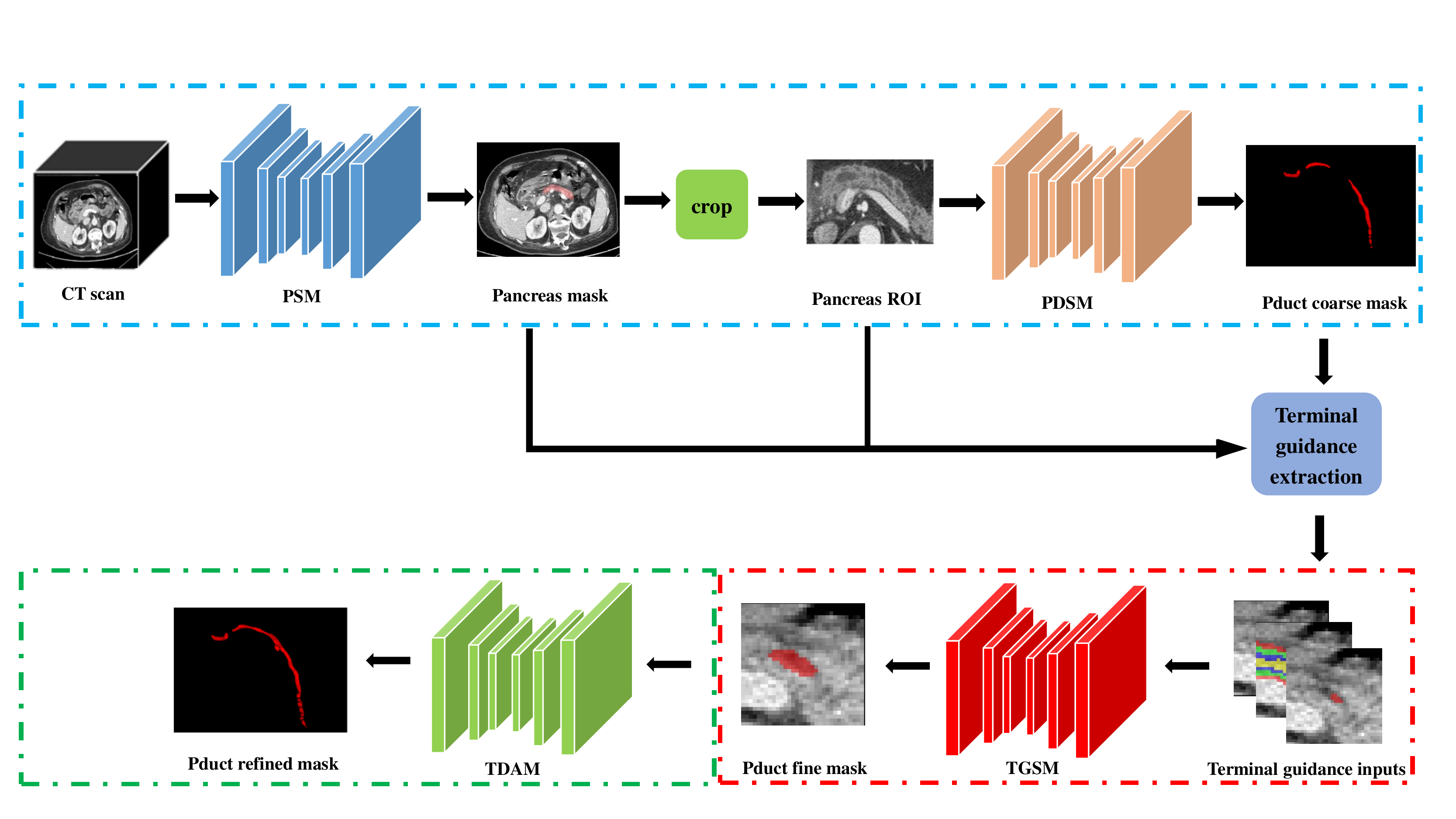}
		}
		\caption{Pipeline of our proposed CTG-Net for dilated pancreatic duct segmentation. The dashed \textcolor{blue}{blue}, \textcolor{red}{red} and \textcolor{green}{green} rectangles denote the coarse, fine and refine stages, respectively. Pduct denotes pancreatic duct.}\label{pipeline}
	\end{figure*}

As shown in Figure \ref{pipeline}, we propose a Cascaded Terminal Guidance Network model, shortly named CTG-Net, by integrating two U-shape segmentation subnetworks in the coarse stage, our designed Terminal Guidance Segmentation Module (TGSM) in the fine stage and the proposed Terminal Distraction Attention Module (TDAM) in the refine stage. The coarse stage is composed of Pancreas Segmentation Module (PSM) and Pancreatic Duct Segmentation Module (PDSM) which are two U-shape segmentation subnetworks. In our experiments, two nnU-Nets \cite{nnunet} are used in the coarse stage. The CT scan is firstly fed into PSM to get coarse pancreas mask and then cropped to obtain the pancreas RoI as the input to the followed PDSM to get the coarse duct segmentation. In the fine stage, the coarse ducts are skeletonized to extraction the terminal patches, and the designed TGSM concatenates the original CT image, coarse prediction and the distance transform map of the pancreas binary mask to an U-Net with three-channel inputs to get the fine duct segmentation. Last but not least, a pre-trained TDAM composed of another U-Net with two-channel inputs processes the original images and the fine duct segmentation to find the FP and FN regions and refine the results based on the fine duct segmentation.

\subsection{Coarse stage for duct segmentation}
Cascaded strategies have been widely used in medical imaging segmentation especially in small organs and lessions\cite{Shen}. The pancreatic duct occupies a tiny portion of abdominal CT volumes, therefore, in order to concentrate on the pancreas regions where pancreatic ducts exist, we need to generate a RoI related to the pancreas.

We train two nnU-Net models to segment the pancreas and pancreatic ducts, respectively. We use the 3-dimension full resolution training strategy that is randomly cropping a patch with fixed size (In our experiments, the patch size is calculated to be $96 \times 160 \times 160$) as the input to the U-shape segmentation network. The cross-entropy (CE) loss and Dice loss are used with 1:1 weight in the training process. And the sliding window strategy is used for testing.

\subsection{Fine stage with TGSM}
The attention mechanism is widely utilized to enhance the influence of useful information and suppress the useless context in deep learning-based approaches. Various attention techniques are investigated for medical image analysis. For dilated pancreatic duct segmentation, in order to use attention-guided framework to improve the segmentation results, \cite{ShenC} adopted the pancreas mask as the guide information to segment the pancreatic duct. However, when one solely uses the pancreas mask as the input information to the network, which only provides the information of the pancreas location without the structure hints inside the pancreas or the guidance for terminal region of the duct predictions.

\begin{figure}[h]
		\centering
		{
			\includegraphics[width=12cm,]{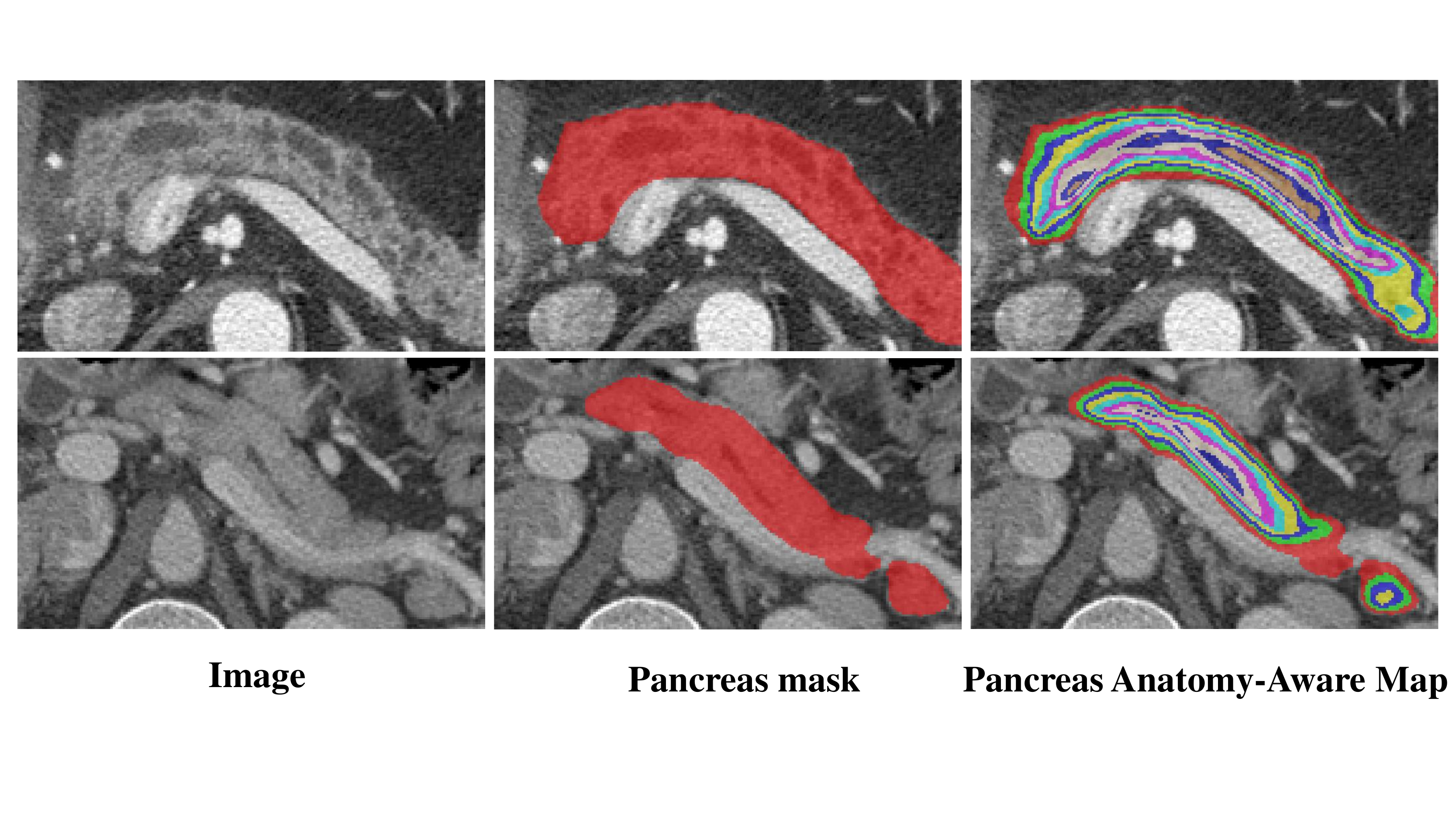}
		}
		\caption{Comparison between the original CT image, pancreas binary mask and our proposed pancreas anatomy-aware (PAA) map calculated from 2 cases in our DPDS dataset. In the PAA map, different colors denote different distances to the boundary of the pancreas binary mask.}\label{paa}
	\end{figure}

To address the problem, we design a terminal guidance segmentation module (TGSM) to focus on segmenting the duct terminals. We firstly extract the terminal regions of the previous predictions by skeletonizing the coarse mask with the method proposed in \cite{Lee}. Secondly, we build the graphs on these skeletons and get the endpoints of these skeletons by neighbor selection. Then, we design an U-shape network with 3-channel inputs to process the output information from the last stage. The terminal CT image and coarse predictions encode the original intensity and feature cues of the targets. As for the third input, instead of the pancreas binary masks, we use the distance transform maps of the pancreas masks which contain more structure hints about the pancreas to improve terminal segmentation performance. The distance transform maps on local terminal RoIs show more global anatomy information inside the pancreas compared with pancreas binary masks. So we term the distance transform map of the pancreas as pancreas anatomy-aware (PAA) map. The PAA map is calculated as follows.

\begin{equation}\label{dist}
	PAA(x,\hat{Y})=
	\left\{
	\begin{array}{lr}
    \underset{y \in \partial \hat{Y}}{\min}d(x,y), &x \in \hat{Y} \\
	0, &otherwise
	\end{array}
	\right.
	\end{equation}
where $x$ denotes the location of voxel in the CT image, $\hat{Y}$ denotes the predicted pancreas mask,  $\partial \hat{Y}$ denotes the boundary of $\hat{Y}$, and the function $d(x,y)$ calculates the Euclidean distance between point $x$ and $y$. Figure \ref{paa} shows the difference between the general pancreas mask and our proposed PAA map.

\subsection{Refine stage with TDAM}

\begin{figure}[h]
		\centering
		{
			\includegraphics[width=12cm,]{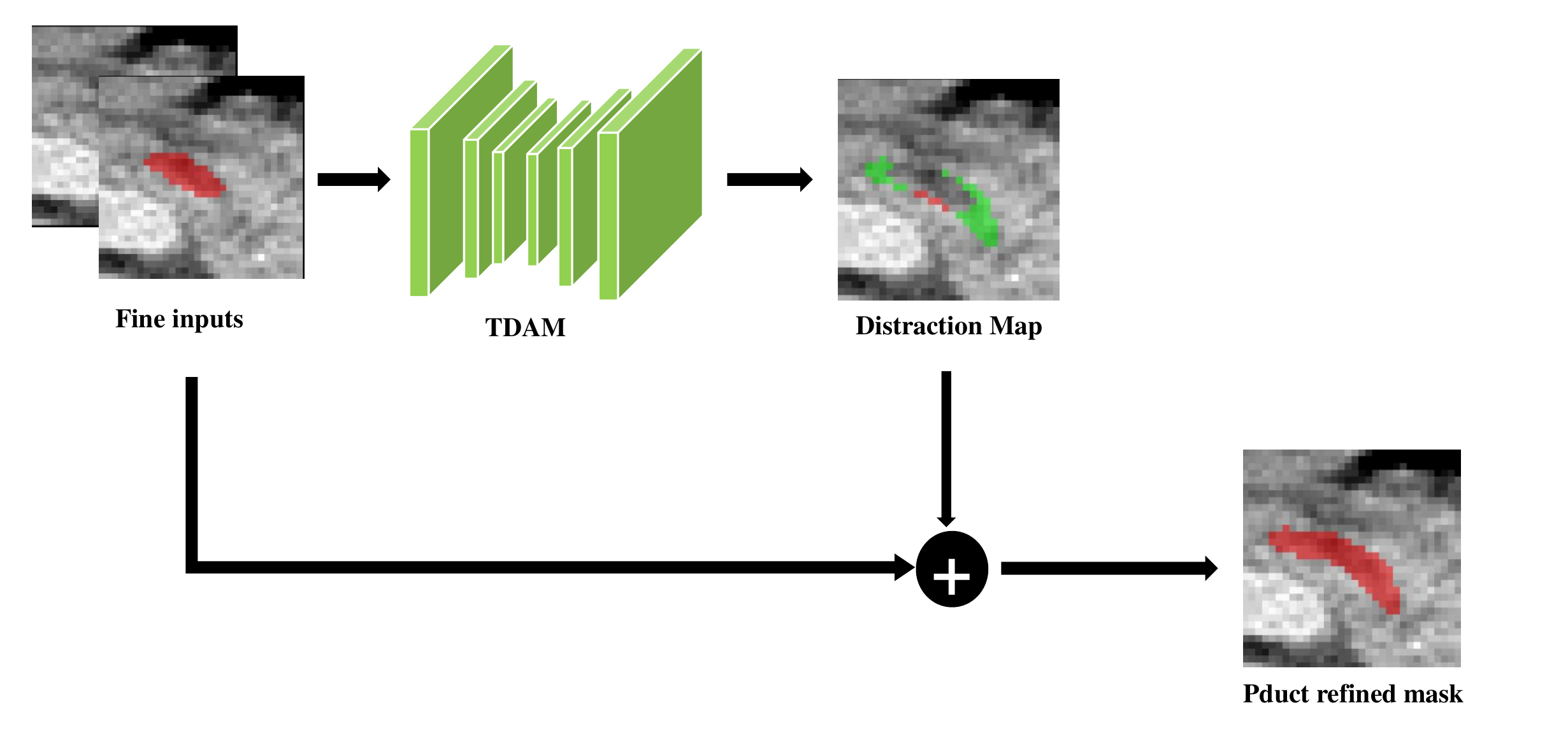}
		}
		\caption{Details of our proposed Terminal Distraction Attention Module. In the distraction map, green voxels denote the predicted FN regions while red voxels denote the predicted FP regions.}\label{TDAM}
	\end{figure}

In the refine stage, we propose a terminal distraction attention module (TDAM) shown in Figure \ref{TDAM} to explicitly learn the distribution of the distraction regions such as bile ducts, tumors and other low contrast regions. It is designed to predict the false positive and false negative regions from previous predictions as possible. During training phase, an U-shape network is trained with two-channel inputs (CT image and the fine prediction) to predict the distraction regions calculated by the difference between ground truth and the previous prediction. At the testing phase, the TDAM outputs the predicted the possible false positive and false negative regions for the fine predictions, then it adds the FN regions and removes the FP regions to get the final segmentation results.

\section{Experiments}
\subsection{Dataset and evaluation}

In this work, we evaluate our proposed methods on our dilated pancreatic duct segmentation (DPDS) dataset collected from Nanjing Drum Tower Hospital. It consists of 150 patients with surgical pathology-confirmed pancreatic tumors (103 PDACs, 32 IPMNs, 7 SCNs, 5 SPTs, 3 MCNs) and dilated pancreatic ducts. Each patient has contrast-enhanced abdominal CT scan and each CT volume contains 165-789 slices with size $512 \times 512$ pixels. The manual annotations of dilated pancreatic duct were performed by two experienced pancreatic imaging radiologists. The dilated pancreatic duct voxels labeled by the experts range from 238 to 49490. All experiments are performed using nested 5-fold cross-validation.

We employ the Dice-S{\o}rensen similarity coefficient (DSC) and Hausdorff distance (HD) as the segmentation metrics which can be calculated as follows.
\begin{equation}\label{dsc}
  DSC=\frac{2TP}{2TP+FP+FN}
\end{equation}

\begin{equation}\label{hd1}
  h(X,Y)=\underset{x \in X}{\max} \underset{y \in Y}{\min}d(x,y)
\end{equation}

\begin{equation}\label{hd2}
  HD(X,Y)=max(h(X,Y),h(Y,X))
\end{equation}
where $TP$, $FP$ and $FN$ in Eq. (2) denote the number of true positive, false positive and false negative voxels in the prediction. $X$ and $Y$ in Eq. (3)-(4) denote the prediction mask and ground truth mask, respectively. In addition to the above two traditional metrics, we propose two new segmentation metrics for distractions. We define the false positive segmentation rate (FPSR) and false negative segmentation rate (FNSR) to meet $DSC+FPSR+FNSR=1$, which can intuitively reflect the false positive and false negative segmentation errors.
\begin{equation}\label{fpr}
  FPSR=\frac{FP}{2TP+FP+FN}
\end{equation}

\begin{equation}\label{fnr}
  FNSR=\frac{FN}{2TP+FP+FN}
\end{equation}

We employ clDice \cite{cldice} to measure the topology preservation of the duct predictions. Moreover, we propose a new metric developed from clDice, called terminal Dice (tDice), to measure the terminal segmentation performance for targets with tubular structures. We consider two binary masks: the ground truth mask ($V_{G}$) and the predicted segmentation mask ($V_{P})$. $R_{G}$ and $R_{P}$ are two binary 3D masks as the same sizes as $V_{G}$ and $V_{P}$. The values of $R_{G}$ and $R_{P}$ can be calculated as follows.
\begin{equation}\label{dist}
	R_{G}(x)=
	\left\{
	\begin{array}{lr}
    V_{G}(x), &\Vert x-\underset{y \in O_{G}}{\arg\min}d(x,y)\Vert_{\infty}\leq \frac{d}{2} \\
	0, &otherwise
	\end{array}
	\right.
	\end{equation}

\begin{equation}\label{dist}
	R_{P}(x)=
	\left\{
	\begin{array}{lr}
    V_{P}(x), &\Vert x-\underset{y \in O_{P}}{\arg\min}d(x,y)\Vert_{\infty}\leq \frac{d}{2} \\
	0, &otherwise
	\end{array}
	\right.
	\end{equation}
where $R_{G}(x)$ and $R_{P}(x)$ denote the values of $R_{G}$ and $R_{P}$ on location $x$, $O_{G}$ and $O_{P}$ denote the sets of endpoints on the skeletons extracted from the $V_{G}$ and $V_{P}$ respectively, and $d$ denotes the size of the terminal RoI. The infinite norm is defined as $||x||_{\infty}=\underset{1 \leq i \leq N}{\max} |x_{i}|$ where $x_{i}$ is the $i$-th component of $x$ with $N$ dimensions. Then, we calculate the terminal precision (Tprec) and terminal sensitivity (Tsens) as defined bellow.

\begin{equation}\label{tprec}
  Tprec(R_{P},V_{G})=\frac{|R_{P} \cap V_{G}|}{|R_{P}|}
\end{equation}

\begin{equation}\label{tsens}
  Tsens(R_{G},V_{P})=\frac{|R_{G} \cap V_{P}|}{|R_{G}|}
\end{equation}

The measure Tprec is related to the false positive predictions while the measure Tsens reflects the false negative predictions. Since we want to maximize both precision and sensitivity, we define tDice to be the harmonic mean of both the measures:
\begin{equation}\label{tdsc}
  tDice(V_{P},V_{G})=2 \times \frac{Tprec \times Tsens}{Tprec+Tsens}
\end{equation}

\subsection{Segmentation results on DPDS dataset}

We get the segmentation results of the dilated pancreatic duct with 5 types of pancreatic tumors in our DPDS dataset through 5-fold cross-validation experiments. We show the overall results of the DSC, HD, FPSR and FNSR in Table \ref{tabsegresults}. Since there are no annotations for pancreas in our DPDS dataset, we show the pretrained segmentation results of the pancreas from 200 cases in the AbdomenCT-1K dataset \cite{abdct} as a reference. It should be pointed out that it is not our goal to segment the pancreas in our work, the pretrained model for pancreas segmentation is accurate enough to obtain the pancreas RoI for the dilated pancreatic duct segmentation.

\begin{table}[htb]
        \centering
		\caption{Segmentation results for the pancreas and dilated pancreatic duct on our DPDS dataset.}
		\vspace{0pt}	
		\renewcommand\arraystretch{2}
		\setlength{\tabcolsep}{6mm}{}{
			\begin{tabular}{c|c|c|c|c}
				\hline
				 Target              &DSC($\%\uparrow$)      & HD(mm$\downarrow$)     & FPSR($\%\downarrow$)  & FNSR($\%\downarrow$)   \\
				\hline
				Pancreas          &85.61$\pm$9.00     &24.69$\pm$47.69       &6.58$\pm$4.99     &7.82$\pm$9.22      \\
                Pancreatic Duct  &84.17$\pm$9.16     &11.11$\pm$13.71      &8.23$\pm$7.37  &7.60$\pm$5.94   \\
				\hline
		\end{tabular}}
		\label{tabsegresults}
	\end{table}

We also get the segmentation results of the dilated pancreatic ducts for patients with different types of pancreatic tumors in our experiments. We show the results of the tDice, clDice, DSC, FPSR and FNSR in Table \ref{tabtumor}. As we can see, the PDAC predictions have the lowest false positive errors, the IPMN predictions achieve the highest Dice score, and the MCN predictions get the best terminal accuracy and lowest false negative errors.

\begin{table}[htb]
        \centering
		\caption{Comparison of segmentation results for dilated pancreatic ducts with different pancreatic tumors on our DPDS dataset.}
		\vspace{0pt}	
		\renewcommand\arraystretch{2}
		\setlength{\tabcolsep}{4mm}{}{
			\begin{tabular}{c|c|c|c|c|c}
				\hline
				Case  &tDice($\%\uparrow$) &clDice($\%\uparrow$)  & DSC($\%\uparrow$)   &FPSR($\%\downarrow$)  & FNSR($\%\downarrow$)    \\
				\hline
				PDAC  &78.53$\pm$14.19 &85.26$\pm$9.92  &84.92$\pm$7.85 &\textbf{7.33$\pm$4.79} &7.75$\pm$6.53\\
                IPMN  & 72.67$\pm$14.61 &\textbf{87.83$\pm$5.69}  &\textbf{85.41$\pm$7.80} &8.08$\pm$7.52 &6.51$\pm$3.60\\
                MCN&\textbf{85.53$\pm$12.70}&86.77$\pm$5.06  &73.90$\pm$15.57&21.03$\pm$17.58&\textbf{5.07$\pm$3.18}\\
                SCN&55.30$\pm$27.24&76.43$\pm$12.56 &72.81$\pm$17.08&16.20$\pm$16.49&10.99$\pm$5.16\\
                SPT&76.97$\pm$11.50&81.98$\pm$6.89   &82.80$\pm$6.16&8.90$\pm$4.71&8.30$\pm$4.87 \\
				\hline
		\end{tabular}}
		\label{tabtumor}
	\end{table}

\subsection{Ablation study}
\begin{figure}[htb]
		\centering
		{
			\includegraphics[width=15cm,]{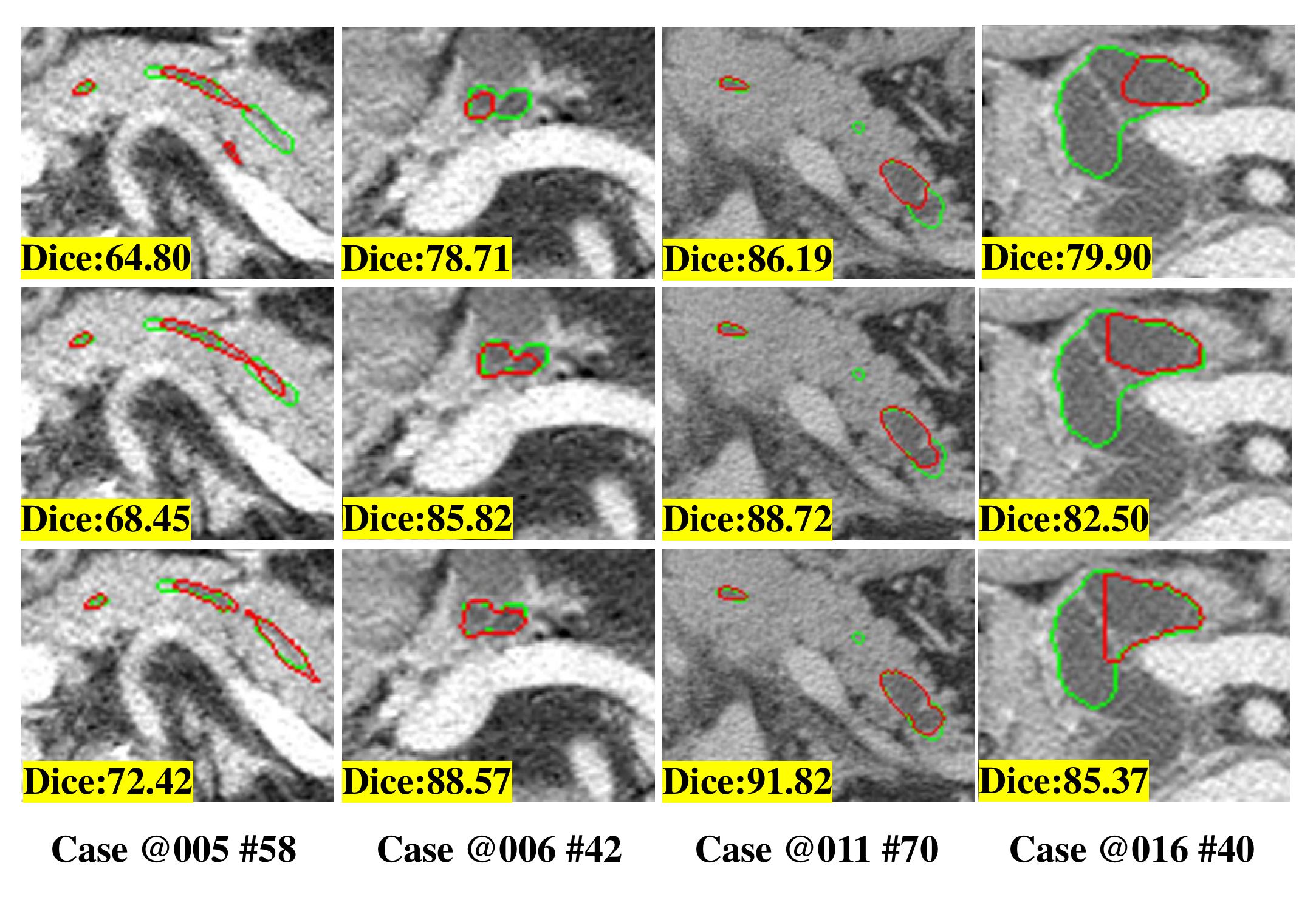}
		}
		\caption{Visual comparison results of the ablation study in the axial plane. The green and red contours denote the ground truth and the predicted segmentation, respectively. Dice score of each case is written in the bottom left of each image. \textbf{Top:} coarse segmentation. \textbf{Middle:} fine segmentation. \textbf{Bottom:} refined segmentation. @ and \# are followed by patient No. and slice No., respectively.}\label{abvistwod}
	\end{figure}

\begin{figure}[htb]
		\centering
		{
			\includegraphics[width=15cm,]{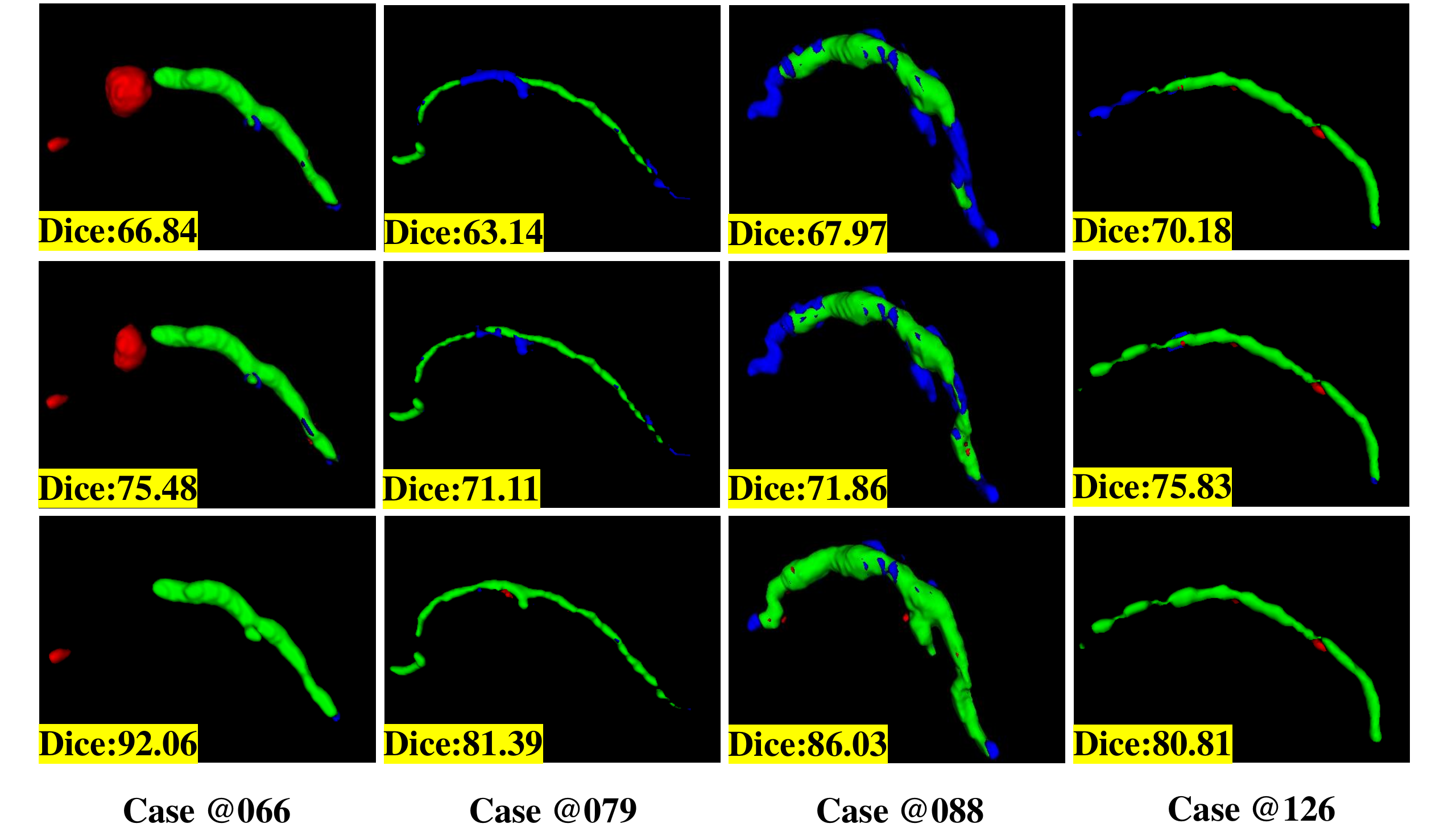}
		}
		\caption{The comparison results of the ablation study in the 3D render. The green, red and blue voxels denote the true positive, false positive and false negative segmentation, respectively. Dice score of each case is written in the bottom left of each image. \textbf{Top:} coarse segmentation. \textbf{Middle:} fine segmentation. \textbf{Bottom:} refined segmentation. }\label{abvisthrd}
	\end{figure}

To verify our improvement of coarse-to-fine and fine-to-refine, we show the ablation study results in Table \ref{tabab}. We can see that our TGSM indeed improves the segmentation results in terms of tDice, clDice, DSC, FPSR and FNSR metrics comparing to the general cascaded strategy used in the coarse stage. Additionally, our TDAM in the refine stage shows its effectiveness to capture terminal distraction regions and further improve the segmentation accuracy. We show the visualization comparison of the ablation study results in Figure \ref{abvistwod} (axial plane) and Figure \ref{abvisthrd} (3D render).

\begin{table}[htb]
        \centering
		\caption{Ablation study for dilated pancreatic duct segmentation on our proposed DPDS dataset.}
		\vspace{0pt}	
		\renewcommand\arraystretch{2}
		\setlength{\tabcolsep}{4mm}{}{
			\begin{tabular}{c|c|c|c|c|c}
				\hline
			    Stage &tDice($\%\uparrow$) &clDice($\%\uparrow$) &DSC($\%\uparrow$)&FPSR($\%\downarrow$) &FNSR($\%\downarrow$)\\
				\hline
				Coarse & 67.25$\pm$16.50 &80.46$\pm$11.62 &78.82$\pm$12.18 & 11.09$\pm$10.50 &10.09$\pm$8.27   \\
				Fine   & 69.99$\pm$16.23 &81.63$\pm$10.64   & 80.18$\pm$11.21 & 9.97$\pm$9.59 & 9.85$\pm$8.26   \\
                Refine & \textbf{76.29$\pm$15.96} &\textbf{85.32$\pm$9.45}  & \textbf{84.17$\pm$9.16}  & \textbf{8.23$\pm$7.37}          &\textbf{7.60$\pm$5.94}         \\
				\hline
		\end{tabular}}
		\label{tabab}
	\end{table}

We further show the improvement of our proposed PAA guidance for terminal segmentation comparing to general methods. We term the methods that feed origin RoI images to the segmentation subnetwork in the fine stage as fine-V1, origin RoI images along with coarse duct predictions as fine-V2, origin RoI images, coarse duct predictions and binary pancreas masks as fine-V3, origin RoI images, coarse duct predictions and PAA maps as fine-V4 (our proposed method), respectively. The segmentation results are shown in Table \ref{tabab2}. We can see that coarse predictions containing feature cues help to improve the terminal segmentation. Additionally, the PAA map shows better performance than binary pancreas mask in terms of raising terminal segmentation accuracy, topology preservation and reducing false positive errors.

\begin{table}[htb]
        \centering
		\caption{Ablation study for the fine stage.}
		\vspace{0pt}	
		\renewcommand\arraystretch{2}
		\setlength{\tabcolsep}{3mm}{}{
			\begin{tabular}{c|c|c|c|c|c}
				\hline
			    Methods&tDice($\%\uparrow$)&clDice($\%\uparrow$)&DSC($\%\uparrow$)&FPSR($\%\downarrow$) &FNSR($\%\downarrow$)\\
				\hline
				Fine-V1 & 69.39$\pm$16.56 &80.61$\pm$10.89  &79.44$\pm$11.30 & \textbf{9.85$\pm$9.85} &10.71$\pm$7.95     \\
				Fine-V2 & 69.46$\pm$16.82 &81.19$\pm$11.06   &79.86$\pm$11.14  & 10.24$\pm$9.36 & 9.89$\pm$8.44        \\
                Fine-V3 &69.83$\pm$17.42  &81.39$\pm$10.76     & 80.06$\pm$11.38  & 10.12$\pm$9.88 & \textbf{9.82$\pm$8.16}        \\
                Fine-V4     & \textbf{69.99$\pm$16.23} &\textbf{81.63$\pm$10.64}    & \textbf{80.18$\pm$11.21}    & 9.97$\pm$9.59    &9.85$\pm$8.26         \\
				\hline
		\end{tabular}}
		\label{tabab2}
	\end{table}

\subsection{Comparison to the SOTA methods}

We compare our CTG-Net with other SOTA methods for medical segmentation on our DPDS dataset. Recently, Transformer \cite{transformer}, a sequence-to-sequence prediction framework, has been considered as an alternative architecture, and has achieved competitive performance on many computer vision tasks, like image recognition \cite{vit}, semantic segmentation \cite{swin}, object detection \cite{carion} and low-level vision \cite{parmar}.  CoTr \cite{cotr} and UNETR \cite{unetr} are the combination of convolutional neural network and transformer for computer vision and achieve SOTA in medical image segmentation challenges. We compare these two SOTA methods and the nnU-Net \cite{nnunet} with our proposed CTG-Net on our DPDS dataset for dilated pancreatic duct segmentation. All the SOTA methods use the same cascaded strategy of segmenting the pancreas firstly and segmenting duct secondly. All the experiments follow the same 5-fold cross-validation setting.

\begin{figure}[htb]
		\centering
		{
			\includegraphics[width=15cm,]{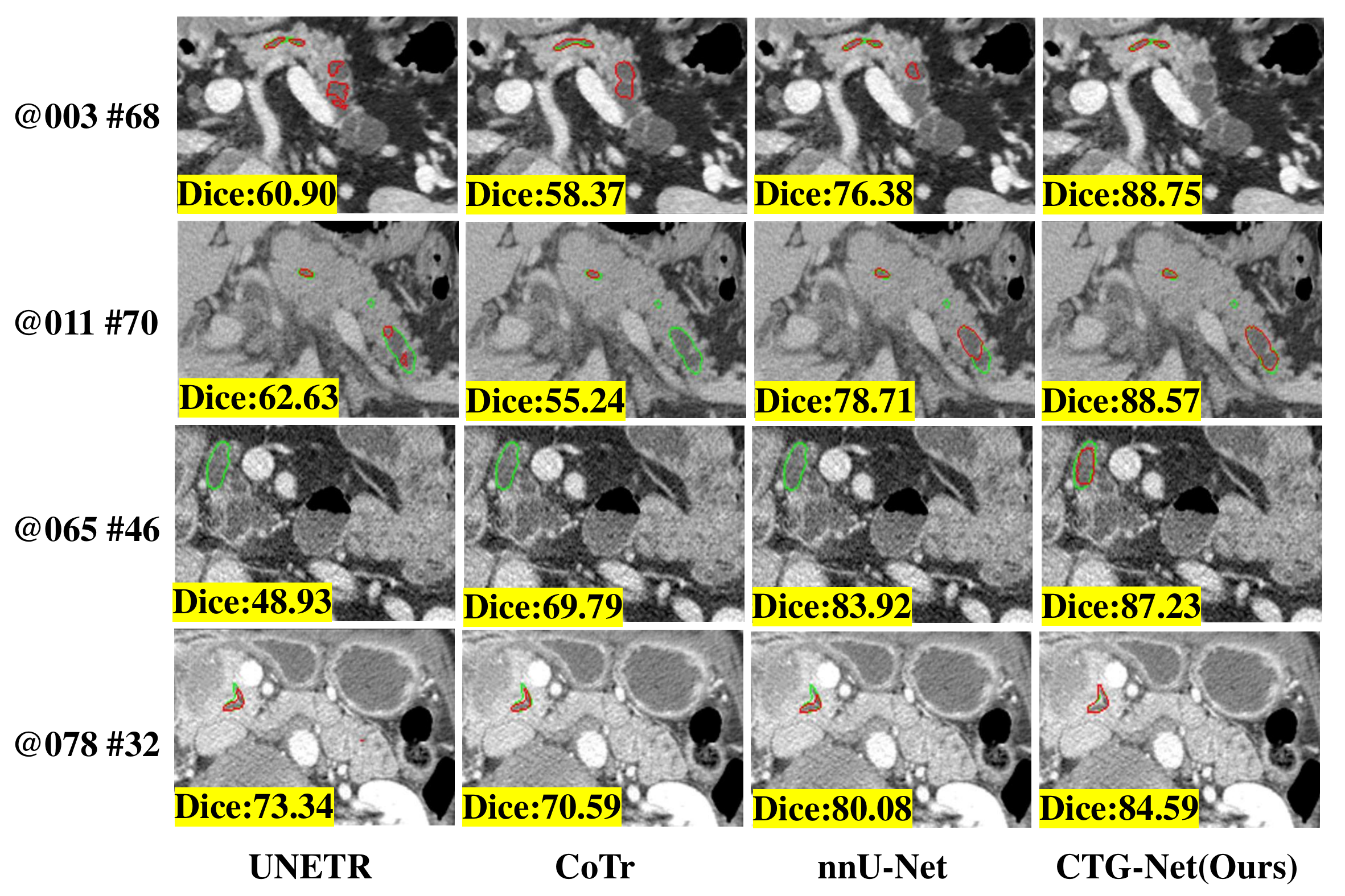}
		}
		\caption{Visual comparison results between the SOTA methods and our CTG-Net in the axial plane. The green and red contours denote the ground truth and the predicted segmentation, respectively. Dice score of each case is written in the bottom left of each image. @ and \# are followed by patient No. and slice No., respectively.}\label{cpvistwod}
	\end{figure}

\begin{figure}[htb]
		\centering
		{
			\includegraphics[width=15cm,]{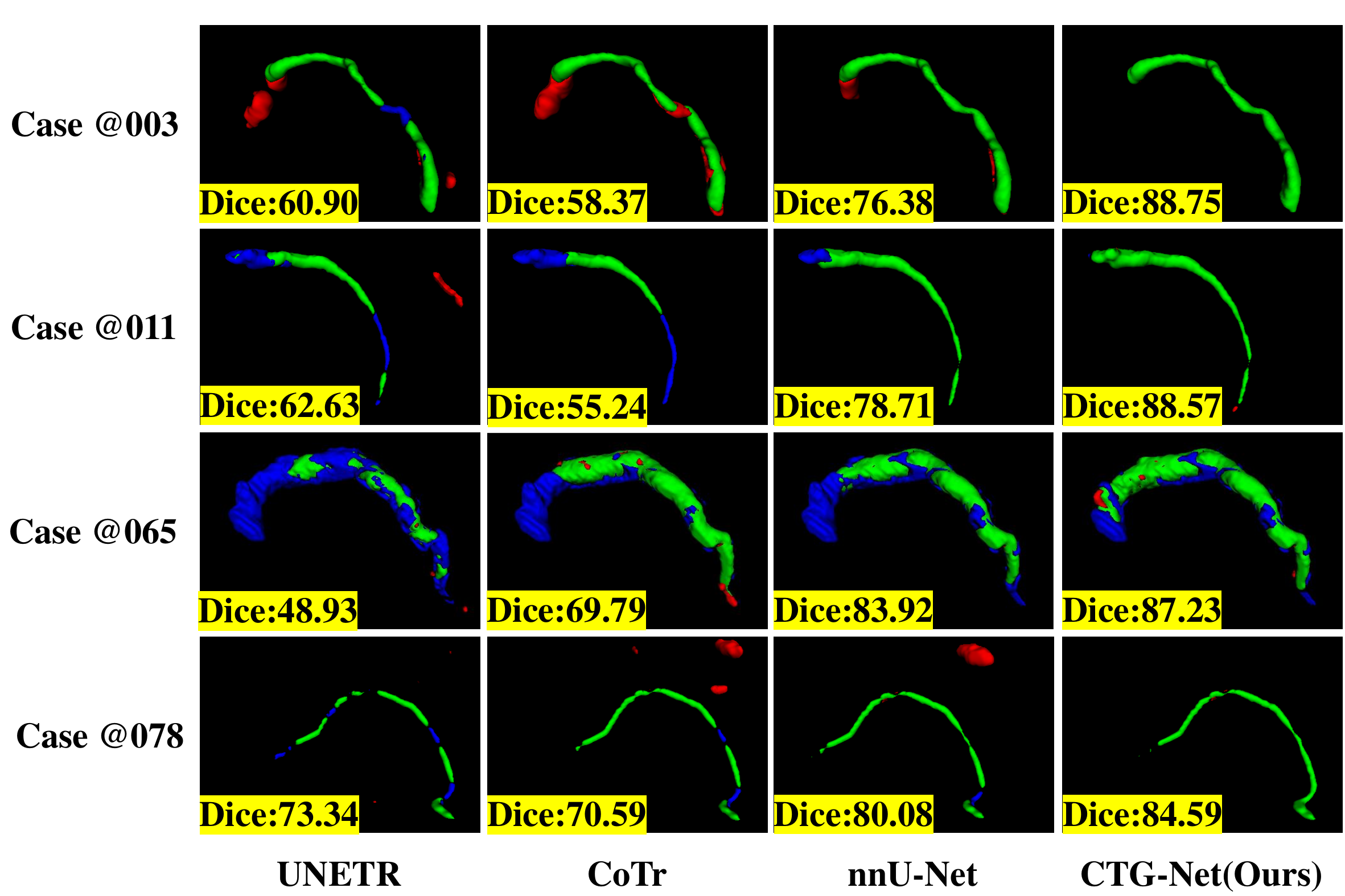}
		}
		\caption{Visual comparison results between the SOTA methods and our CTG-Net in the 3D render. The green, red and blue voxels denote the true positive, false positive and false negative segmentation, respectively. Dice score of each case is written in the bottom left of each image. }\label{cpvisthrd}
	\end{figure}

The segmentation results are shown in Table \ref{tabcp} where we can see that our CTG-Net achieves the best performance in terms of tDice, clDice, DSC, FPSR and FNSR metrics. Additionally, the transformer-combined methods do not show their superiority to the CNN-based methods, which may be caused by the limited scale of our collected dataset. Figure \ref{cpvistwod} and Figure \ref{cpvisthrd} show the visual comparison of axial plane and 3D render results, respectively.
\begin{table}[htb]
        \centering
		\caption{Comparison to the SOTA methods.}
		\vspace{0pt}	
		\renewcommand\arraystretch{2}
		\setlength{\tabcolsep}{3mm}{}{
			\begin{tabular}{c|c|c|c|c|c}
				\hline
			    Methods&tDice($\%\uparrow$)&clDice($\%\uparrow$)&DSC($\%\uparrow$)&FPSR($\%\downarrow$) &FNSR($\%\downarrow$)\\
				\hline
                nnU-Net & 67.25$\pm$16.50 &80.46$\pm$11.62  &78.82$\pm$12.18 & 11.09$\pm$10.50 &10.09$\pm$8.27   \\
				CoTr & 64.02$\pm$18.36 &75.76$\pm$18.05   &74.28$\pm$18.55 &9.01$\pm$9.58  &16.72$\pm$19.03  \\
                UNETR & 56.90$\pm$16.04&72.13$\pm$13.50    &71.57$\pm$13.81 &11.54$\pm$9.96  &16.90$\pm$11.97  \\
                \hline
                \textbf{CTG-Net}   &\textbf{76.29$\pm$15.96} &\textbf{85.32$\pm$9.45}  &\textbf{84.17$\pm$9.16}  &\textbf{8.23$\pm$7.37}       &\textbf{7.60$\pm$5.94} \\
				\hline
		\end{tabular}}
		\label{tabcp}
	\end{table}

\subsection{Comparison to the previous techniques on dilated pancreatic duct segmentation}
We also compare with the previous techniques for dilated pancreatic duct segmentation in Table \ref{tabsota}. The performance of anatomy-aware attention with the distance transform for the terminal guidance segmentation we have developed outperforms the previous works \cite{Shen,ShenC}. This may be due to the fact that these works only use the pancreas/non-pancreas information, which might ignore the anatomy structure information of the pancreas with pancreatic duct. And we also beat the \cite{Xia,Zhou} because pancreatic duct is a extremely tiny component that only occupies a small part of an image, the coarse-to-fine framework shows its superiority. Additionally, all the above works do not concentrate on the terminal errors where we improve a lot. Although we can not make a direct comparison to these techniques, our methods achieve the highest reported DSC for pancreatic duct segmentation on our dataset with moderate quantity and the most tumor types. Moreover, our performance is based on single phase CT volumes. Multi-phase CT volumes can help the model to learn more information about the dilated pancreatic ducts.

\begin{table}[htb]
        \centering
		\caption{Comparison to previous dilated pancreatic duct segmentation methods.}
		\vspace{0pt}	
		\renewcommand\arraystretch{2}
		\setlength{\tabcolsep}{4mm}{}{
			\begin{tabular}{c|c|c|c|c}
				\hline
				Methods          &Phase          & Data   & Tumor types  & DSC(\%)  \\
				\hline
				SE-Dense Unet\cite{Shen}   & Single  & 30      & 1           &  49.87     \\
				MPA Net\cite{ShenC}  & Single        & 30          & 1      &  54.16     \\
                HPN\cite{Zhou}   & Multi         & 239         & 1           &  56.77     \\
                Alignment Ensemble\cite{Xia} & Multi  & 239    & 1  &  64.38     \\
                \hline
                \textbf{CTG-Net (Ours)} &Single  &150     &\textbf{5} & \textbf{84.17}    \\
				\hline
		\end{tabular}}
		\label{tabsota}
	\end{table}

\section{Discussions}
We have successfully evaluated the proposed CTG-Net on our DPDS dataset which includes dilated pancreatic ducts with different types of tumors. In this section, we give the prediction results of TDAM in our experiments. Table \ref{tabdis} shows the segmentation results of FP and FN regions predicted by TDAM. The DSC, HD, FPSR and FNSR are calculated on the terminal regions. Because of the tiny size of FP and FN regions, minor errors can make great drop to the segmentation metrics, which makes the Dice score of the TDAM's predictions within terminal RoIs not high. Nevertheless, since the predictions are only focus the terminal regions, over 45\% terminal distraction accuracy can still make a big improvement for the final results of the whole CTG-Net framework, which shows the excellent robustness of our proposed method.

\begin{table}[htb]
        \centering
		\caption{Segmentation results of TDAM.}
		\vspace{0pt}	
		\renewcommand\arraystretch{2}
		\setlength{\tabcolsep}{3mm}{}{
			\begin{tabular}{c|c|c|c|c}
				\hline
				 Target     &DSC($\%\uparrow$)   &HD(mm$\downarrow$) & FPSR($\%\downarrow$)  & FNSR($\%\downarrow$)   \\
				\hline
				False Positive      &\textbf{54.09$\pm$26.78}     &8.98$\pm$5.89       &\textbf{13.08$\pm$12.89}  &\textbf{23.77$\pm$18.26}      \\
                False Negative      &46.84$\pm$27.61     &\textbf{7.50$\pm$4.90}      &15.68$\pm$16.90&28.65$\pm$24.05   \\
				\hline
		\end{tabular}}
		\label{tabdis}
	\end{table}

We use the nnU-Net as the U-shape network in the fine and refine stage of our CTG-Net. However, it can be replaced by any other segmentation network. We set the size $d$ of the terminal RoI in our experiments to 32 voxels which is an optimal hyperparameter after considering the sizes of the pancreas and pancreatic ducts. More experiments about adaptive parameter selection of the terminal size can be implemented in our subsequent studies.

Our proposed method follows the intuitive and effective coarse-fine-refine strategy for medical segmentation. In fact, all the 3 stages containing 4 networks in our CTG-Net can be trained in an end-to-end manner to find the global optimal parameters for the whole framework, which is also studied in our further work. Moreover, the tDice metric proposed in this work can be used as a loss function to train a segmentation network for improving the terminal segmentation accuracy.

\section{Conclusions}
In this paper, we propose a cascaded framework driven by terminal guidance mechanism for dilated pancreatic duct segmentation. Our approach is inspired not only by the effective coarse-to-fine strategy for lesion segmentation, but also the insufficient segmentation of the targets with tubular structures by traditional methods. We employ the general cascaded strategy as the coarse predictions which are fed into the proposed terminal guidance segmentation module learning local intensity, feature cues and anatomy structure information to get the fine results on the terminal regions. Finally, the model predicts the distraction regions learned by the proposed terminal distraction attention module to refine the final results. In our experiments, we achieve the best segmentation performance comparing not only the SOTA methods based on CNN and transformers, but also previous techniques for dilated pancreatic duct segmentation. Additionally, our experimental dataset contains 5 types of pancreatic tumor patients with dilated pancreatic ducts, which is not studied in previous works. Our method can not only apply to other medical segmentation tasks such as organ-specific tumors, but also the segmentation challenges for targets with tubular structures.

\section*{Acknowledgement}
This work is supported by Ministry of Science and Technology of the People's Republic of China(No. 2020YFA0713800), National Natural Science Foundation of China(No. 11971229, 12001273) and Postgraduate Research \& Practice Innovation Program of Jiangsu Province(No. KYCX22\_0082).

\end{document}